\documentclass[5pt]{elsarticle}
\let\today\relax
\makeatletter
\def\ps@pprintTitle{%
	\let\@oddhead\@empty
	\let\@evenhead\@empty
	\def\@oddfoot{\footnotesize\itshape
		{Accepted in the Journal of Computational Science} \hfill\today}%
	\let\@evenfoot\@oddfoot
}
\makeatother

\usepackage{amsmath}
\usepackage[table,xcdraw]{xcolor}
\usepackage[italian]{varioref}
\usepackage[utf8]{inputenc}
\usepackage[nottoc]{tocbibind}
\usepackage{hhline}
\usepackage{diagbox}

\usepackage{lineno}
\modulolinenumbers[50]

\usepackage{dcolumn}
\usepackage{bm}
\usepackage[bf, font=small]{caption}
\usepackage{tabularx,ragged2e,booktabs}

\usepackage{algorithm}
\usepackage{algpseudocode}

\usepackage{mathrsfs}
\usepackage{forest,adjustbox}
\makeatletter
\def\BState{\State\hskip-\ALG@thistlm}
\makeatother

\usepackage{natbib}
\usepackage[italian, english]{babel}
\usepackage{helvet}
\usepackage{mathtools} 

\usepackage{amssymb}
\usepackage{amsfonts}
\usepackage{numprint}
\usepackage{verbatim}
\usepackage{graphicx}
\usepackage{eucal}
\usepackage{rotating}
\usepackage[pdftex]{lscape}
\usepackage{longtable}

\usepackage{color}
\usepackage{bbm}
\usepackage{dsfont}
\makeatletter
\def\BState{\State\hskip-\ALG@thistlm}
\makeatother
\usepackage{tikz}
\usetikzlibrary{3d}
\usetikzlibrary{patterns}
\usetikzlibrary{calc}
\usetikzlibrary{arrows}

\usepackage{pgfplots}
\usepackage{pgfplotstable}
\usetikzlibrary{decorations.pathreplacing}
\makeatletter
\newcommand{\gettikzxy}[3]{%
	\tikz@scan@one@point\pgfutil@firstofone#1\relax
	\edef#2{\the\pgf@x}%
	\edef#3{\the\pgf@y}%
}
\makeatother

\usepackage{multirow}
\usepackage{etoolbox}
\usepackage{hyperref}

\newcommand{\M}[2][]{{\bm{#1\mathbf{\MakeUppercase{#2}}}}} 
\newcommand{\Mn}[3][]{{\bm{#1\mathbf{\MakeUppercase{#2}}}}^{(#3)}} 
\newcommand{\T}[2][]{\boldsymbol{#1\mathcal{\MakeUppercase{#2}}}} 
\newcommand{\Mz}[3][]{\M[#1]{#2}_{(#3)}}

\DeclareMathOperator*{\argmin}{arg\,min}

\newcommand{\tikzcircle}[2][black,fill=black]{\tikz[baseline=-0.5ex]\draw[#1,radius=#2] (0,0) circle ;}
\usepackage{subcaption}
\usepackage{fancyhdr}
\begin{document}
	\begin{frontmatter}	
	\title{Predicting Multidimensional Data via Tensor Learning}
		
\author[1]{Giuseppe Brandi\corref{cor1}}
\ead{giuseppe.brandi@kcl.ac.uk}
\author[1,2,3]{T. Di Matteo}

\cortext[cor1]{Corresponding author}
\address[1]{Department of Mathematics, King's College London, The Strand, London, WC2R 2LS, UK}
\address[2]{Complexity Science Hub Vienna, Josefstaedter Strasse 39, A 1080 Vienna, Austria}
\address[3]{Centro Ricerche Enrico Fermi, Via Panisperna 89 A, 00184 Rome, Italy}

\begin{abstract}
			The analysis of multidimensional data is becoming a more and more relevant topic in statistical and machine learning research. Given their complexity, such data objects are usually reshaped into matrices or vectors and then analysed. However, this methodology presents several drawbacks.
			First of all, it destroys the intrinsic interconnections among datapoints in the multidimensional space and, secondly, the number of parameters to be estimated in a model increases exponentially. We develop a model that overcomes such drawbacks. In particular, in this paper, we propose a parsimonious tensor regression model that retains the intrinsic multidimensional structure of the dataset. Tucker structure is employed to achieve parsimony and a shrinkage penalization is introduced to deal with over-fitting and collinearity. To estimate the model parameters, an Alternating Least Squares algorithm is developed. In order to validate the model performance and robustness, a simulation exercise is produced. Moreover, we perform an empirical analysis that highlight the forecasting power of the model with respect to benchmark models. This is achieved by implementing an autoregressive specification on the Foursquares spatio-temporal dataset together with a macroeconomic panel dataset. Overall, the proposed model is able to outperform benchmark models present in the forecasting literature. 
\end{abstract}



\end{frontmatter}	
	
	\section{Introduction}
		Big Data and multidimensional data are becoming a relevant topic in statistical and machine learning research \cite{anandkumar2014tensor,cichocki2014era,romera2013multilinear,Brandiphd}. Working with such huge and complex objects in order to extract information is difficult for numerous reasons \cite{acar2009future,tumminello2005tool}. Among all, there are the dimension of the dataset and the sophistication of the model necessary to handle the data \cite{acar2009future}. In some cases, in order to analyse huge datasets, the procedure is to reduce its dimensionality or to use approximate methods. The issue is even more pronounced when we deal with multidimensional data. Multidimensional data are generally referred to datasets characterized by more than two dimensions, e.g. 3D images, panel data (individuals $\times$ variables  $\times$ time  $\times$ location) or higher order ($>2$) multivariate portfolio moments \cite{acar2009future,brandi2019unveil}. A na\"{i}ve approach to model such complex-structured datasets is to reshape them into vectors or matrices and then to analyse these lower-dimensional objects through standard statistical techniques. However, this manipulation of the original datasets has some drawbacks. First of all, it destroys the intrinsic interconnections between the data points in the multidimensional space \cite{anandkumar2014tensor, Brandiphd,billio2018bayesian,lock2017,kolda2009tensor}. Secondly, the number of parameters to be estimated increases exponentially \cite{acar2009future,bro1998multi,smilde2005multi,zhang2017matrix}. As far as concern the first drawback, neglecting the interconnections between dimensions makes the models' predictions less accurate and difficult to interpret as we would be treating some strongly interconnected components of the real system as independent. With respect to the second limitation, the reshaped dataset would have a much higher number of rows or columns which in turns will end up in a very complex model or, to be able to estimate it, to its oversimplification. \par
		To overcome these limitations, we propose a model that doesn't require the simplification of the matrix/vector form, yet being parsimonious. Indeed, in this paper we construct a tensor variate regression model that is able to exploit the intrinsic multidimensional structure of the dataset, in the same spirit as in \cite{Brandiphd, billio2018bayesian, lock2017}. In particular, we build a Tensor regression with Tucker structured coefficient. The use of the Tucker structure in the learning model has two main advantages. The first one is the reduction of the complexity of the regression model while the second one is, in the spirit of reduced rank regression, the exploitation of the common information shared by the modes of the tensors \cite{acar2009future,izenman1975reduced,mukherjee2011}.
		In order to handle data collinearity, we also build a Ridge (Tikhonov) regularized version \cite{tikhonov1943stability,tikhonovsolutions,kennedy2003guide,arcucci2017decomposition} of the model which can be very useful in some research fields, e.g. image recognition. Finally, since the estimation of the model's parameters is not available in closed form all at once, we propose an estimation algorithm based on the Alternating Least Squares (ALS), which solves iteratively least squares sub-problems of the full model \cite{kroonenberg1980principal,acar2009future}. As this procedure solves the model learning problem by using sections of the full dataset, it is indeed a powerful tool to apply in big data environments and has become the workhorse for multilinear optimization \cite{Brandiphd, acar2009future}.  To test the reliability of the model and its robustness, we have performed two simulation experiments. The first one is devoted to the estimation of a structural coefficients (e.g. image) while the second is dedicated to the data setting in which collinearity within dimensions is present. Finally, we have performed an empirical analysis by studying the forecasting power of the tensor regression model in a multidimensional time series domain. To do so, we run an autoregressive specification on the Foursquares spatio-temporal dataset and a macroeconomic panel dataset. We find that, on the overall, the proposed model is able to outperform existing models present in forecasting literature. \par
	The paper is structured as follows. Section \ref{sec_np} is related to the notation and operation on tensors. Section \ref{sec_tr} introduces the tensor regression and the proposed estimation procedure. Section \ref{sec_ss} is devoted to the simulation study while Section \ref{sec_ea} to the empirical application with the results. Section \ref{sec_c} concludes. 
	
	\section{Notation and preliminaries}\label{sec_np}
	
	Tensors are generalization of scalars, vectors and matrices and their \textit{order} defines them \cite{acar2009future,kolda2009tensor,kroonenberg1980principal}. The order of a tensor, that is the number of dimensions that characterises it, is also referred to as ways or modes. Scalars are $0$-th order tensors, vectors are $1$-st order tensors and matrices are $2$-nd order tensors. Whatever has more than two dimensions is referred to as higher order tensor or just as tensor. Throughout this work the notation will follow the standard notation introduced in \cite{kolda2009tensor}: $x$ is a scalar, $\mathbf{x}$ is a vector, $\mathbf{X}$ is a matrix and $\T{X}$ is a tensor. 
	
	\subsection{Operations on tensors}
	There are different operations which can be performed with tensors through the use of linear and multilinear algebra. The literature on the topic is vast and \cite{anandkumar2014tensor,kolda2009tensor,kolda2006multilinear,bader2006algorithm,de2000best,liu2008hadamard} are only few of the papers focusing on tensor operations. They carefully present the various calculations involving tensors and matrices both formally and with examples. For the sake of conciseness, we present only the tensor operations inherent to this work.
	\subsubsection{Matricization}
	The matricization of a tensor (also called unfolding) is the process of reshaping a tensor into a matrix.
	Take a tensor $\T{X} \in \mathbb{R}^{I_1 \times I_{2}\cdots \times I_N} $. Let the ordered sets
	$\T{R} = {\{r_1,... , r_L\}}$ and $\T{C} = \{c_1,... , c_M\}$ be a partitioning of the modes of the tensor $\T{N} = \{1,... , N\}$. The matricized tensor is defined as:
	
	\begin{equation}
\mathbf{X}_{(\T{R} \times \T{C},I_{\T{N}})} \in \mathbb{R}^{J \times K}
	\end{equation} 
	
	where $I_{\T{N}}$ is the size of the original tensor, $J=\prod_{n \in \T{R} }I_{n}$ and $K=\prod_{n \in \T{C} }I_{n}$. The \emph{$n$-mode matricization} is a special case in which the set $\T{R}$ is a singleton equal to $n$ and $\T{C} = \{1,...,n-1,n+1,..., N\}$. It is defined as $\mathbf{X}_{(\T{R} \times \T{C},I_{\T{N}})}\equiv \mathbf{X}_{(n)}$. In this case the fibers of mode $n$ are aligned as the columns of the resulting matrix. When $\T{R}=\T{N}$ and $\T{C}=\emptyset$, we have the vectorization of the tensor, i.e.: 
	 \[\mathbf{X}_{(\T{R} \times \emptyset, I_{\T{N}})}\equiv\mathbf{X}_{(\T{N} \times \emptyset, I_{\T{N}})}\equiv vec(\T{X}).\]
	 
	\subsubsection{Tensor multiplication}
	As for the matrix case, it is possible to define different tensors multiplications through the use of linear and multilinear algebra  \cite{kolda2009tensor,kolda2006multilinear,bader2006algorithm}. In particular, we define the \emph{$n$-mode product} and the \emph{contracted product}.
	\subparagraph{n-mode product:}
	
	The \emph{$n$-mode product} of the tensor $\T{X} \in \mathbb{R}^{I_1 \times I_{2}\cdots \times I_N}$ with a matrix $\M{V}$ of size $J \times I_n $ is denoted as: 
	\begin{equation}
\T{y}= \T{X}\times_n \M{V}
	\end{equation}
	 and the resulting tensor $\T{y}$ is of size $I_1 \times \cdots \times I_{n-1} \times J
	\times I_{n+1} \times \cdots \times I_N $. %
	
	This can be expressed in terms of matricized tensors as:
	\begin{displaymath}
	\T{Y} = \T{X} \times_n \M{V} 
	\quad \Leftrightarrow \quad 
	\Mz{Y}{n} = \M{V}\Mz{X}{n}.
	\end{displaymath}
	where $\Mz{Y}{n}$ and $\Mz{X}{n}$ are the \emph{$n$-mode matricization} of $\T{Y}$ and $\T{X}$.
	
	\subparagraph{Contracted product:}
	
	The contracted product, also known as tensor contraction, is the tensor multiplication between tensors which extend the matrix product to higher dimension. Let $\T{X} \in \mathbb{R}^{I_1 \times I_{2}\cdots \times I_N}$ and $\T{V} \in \mathbb{R}^{J_1 \times J_{2}\cdots \times J_M}$. The constructed product is conveniently written as:
	\begin{equation}
	\T{Y}=\langle \T{X},\T{V} \rangle_{(\T{I_X};\T{J_V})},
	\end{equation}
	
	where the subscripts $\T{I_x}$ and $\T{J_V}$ are the modes over winch the product is carried out. The dimension of  $\T{I_x}$ and $\T{J_V}$ can be different but the size of the modes over which the contraction is performed must be identical. This can be computed in matricized form as:

	\[\mathbf{X}_{(\T{I_x} \times \T{I_x^\complement},I_{\T{N}})} \mathbf{V}_{(\T{J_v} \times \T{J_v^\complement},J_{\T{M}})}\]
	
	where $\T{I_x^\complement}$ and $\T{J_v^\complement}$ are the set of modes not involved in the product. The case in which  $\T{x}$ and $\T{V}$ have the same dimensions and the contraction is performed over all the modes, defines the inner product and it results in a scalar, i.e.:
		\[
	y=\langle \T{X},\T{V} \rangle.
	\]
	This definition can be used to compute the (squared) Frobenious norm of a tensor, i.e.:
	
	\[
	\|\T{X}\|^2_F=\langle \T{X},\T{X} \rangle.
	\]
  This can be written in matrix form as $\|\T{X}\|^2_F=	\|\Mz{X}{n}\|^2_F$ as presented in \cite{bader2006algorithm}. This shows that the Frobenious norm of a tensor is equivalent to the Frobenious norm of its \emph{$n$-mode matricization}.
	
	\subsubsection{Tensor decomposition}
	Tensors, like matrices, can be decomposed in smaller (in terms of rank) objects \cite{anandkumar2014tensor,acar2009future,kolda2009tensor}. One factorization method employed in multi-way analysis is the Tucker decomposition theorized by Tucker \cite{tucker1964,tucker1966}. It represents an extension of the bilinear factor analysis to the higher dimensional case. It is also referred to as N-mode PCA \cite{kapteyn1986approach}  and Higher-order SVD \cite{de2000best}. Take an $n$-th order tensor $\T{X} \in \mathbb{R}^{I_1 \times I_{2}\cdots \times I_N} $. The Tucker decomposition of $\T{X}$ takes the form of a \textit{n-mode} product, i.e.:			
	\begin{equation}\label{t}
	\T{X} \approx \T{G} \times_1 \Mn{U}{1} \times_2 \Mn{U}{2} \cdots \times_N \Mn{U}{N} = \T{G} \times \{ \Mn{U}{n} \},
	\end{equation}
	where $\Mn{U}{n}$ are the factor matrices and $\T{G}$ is the \textit{core} tensor which is usually of dimension smaller than  $\T{X}$. The Tucker decomposition can be be employed for two main reasons, i.e. factor analysis or dimensionality reduction. The former is performed by analysing the factor matrices $\Mn{U}{n}$ corresponding to each mode of the decomposed tensor. The latter is achieved when $\T{G}$ is of lower dimension than $\T{X}$. Indeed, it is possible to rewrite (approximately) $\T{X}$ with a much lower number of components. In this paper, we use the Tucker structure on the regression coefficient in order to achieve parsimony and to exploit common information across all tensors' dimensions, in the same spirit of the reduce rank regression approach.

	\section{Tensor regression}\label{sec_tr}
	In this section we propose a Tensor regression with Tucker structured coefficient and develop an ALS type of algorithm to estimate the model parameters.  
	Tensor regression can be formulated in different ways: the tensor structure is only in the dependent or the independent variable or it can be in both. The literature related to the first specification is ample \cite{zhou2013,sun2017,li2017,guhaniyogi2017bayesian,li2018} whilst the fully tensor variate regression received attention only recently from the statistics and machine learning communities employing different approaches \cite{Brandiphd,billio2018bayesian, lock2017, hoff2015multilinear,yu2015}. In this work we will extend the tensor regression model in \cite{Brandiphd} to the arbitrary dimensional case and, in same spirit of \cite{lock2017}, we introduce a penalty to take into account the possible collinearity between the variables in the dataset. The fully tensor variate regression models proposed by \cite{billio2018bayesian,lock2017} have the common feature of having a PARAFAC \cite{kolda2009tensor,bro1998multi,bro1997parafac} structure in the tensor coefficient. They use this specification as it makes the Bayesian modeling approach easier. However, this structural form is a limitation since the PARAFAC decomposition imposes the factor matrices to have the same number of factor components. Indeed, the PARAFAC decomposition can be seen as a constrained Tucker decomposition with super-diagonal core tensor and with the same number of factor components along each dimension. The limitation becomes more evident when we deal with dimensionally skewed tensors, for which imposing the same number of components is excessively restrictive. In contrast, the Tucker structure remove this limitation by allowing an arbitrary number of factor components in each mode. The proposed tensor regression is formulated making use of the contracted product and can be expressed as: 
	
	\begin{equation}\label{tr}
	\T{Y}=\T{A} + \langle \T{X},\T{B} \rangle_{(\T{I_x};\T{I_B})}+\T{E},
	\end{equation}
	where $\T{X}\in \mathbb{R}^{N \times  I_1 \times \cdots \times I_N}$ is the regressor tensor, $\T{Y} \in \mathbb{R}^{N \times J_1\times \cdots \times J_M} $ is the response tensor, $\T{E}\in \mathbb{R}^{N \times J_1\times\cdots \times J_M}$ is the error tensor, $\T{A}\in \mathbb{R}^{1 \times J_1\times\cdots \times J_M}$ is the intercept tensor while the slope coefficient tensor we are interested to learn is $\T{B}\in \mathbb{R}^{I_1 \times \cdots \times I_N \times J_1 \times \cdots \times J_M}$. The subscripts $\T{I_x}$ and $\T{J_B}$ are the modes over which the product is carried out. Unfolding $\T{Y}$, $\T{X}$, $\T{E}$ and $\T{A}$ over the first $(n+1)$ modes and $\T{B}$ over the first $(N-n-1)$ modes, we can rewrite the regression (omitting subscripts for the sake of simplicity) as:
	
		\[\mathbf{Y}= \mathbf{A} + \mathbf{X}\mathbf{B} + \mathbf{E}.
		\]
	The regression takes the form of a multivariate regression for which the Least Squares (LS) solution (given sufficient data) of $\mathbf{B}$ is:
	\[
	\mathbf{\widehat{B}}=(\mathbf{X}^{T} \mathbf{X})^{-1} \mathbf{X}^{T}\mathbf{Y}
	\]
	
	and
	
		\[
	\mathbf{\widehat{A}}=\mathbb{E}[\mathbf{Y}]-\mathbb{E}[\mathbf{X}]\mathbf{\widehat{B}}.
	\]
	This equivalent representation of Eq. \ref{tr} shows the relationship between multivariate regression and tensor regression. Indeed, Eq. \ref{tr} reduces to multivariate regression if $\T{Y}$ and $\T{X}$ are $2$-nd order tensors (matrices).	However, the $\T{B}$ coefficient is high dimensional. In order to resolve the issue and to enhance parsimony, we impose a Tucker structure on $\T{B}$ such that it is possible to recover the original $\T{B}$ by a combination of smaller objects, i.e.:
	
	\begin{equation}\label{tb}
	\T{B} \approx \T{G} \times_1 \Mn{U}{1} \cdots \times_N \Mn{U}{N} \times_{N+1} \Mn{V}{1} \times \cdots \times_{N+M} \Mn{V}{M}
	\end{equation}
	where $\T{G}$ is the core tensor which drives the complexity of the reconstruction, $\Mn{U}{n}$ are the factor matrices related to the input modes while $\Mn{V}{m}$ are the factor matrices related to the output modes. This specification has two advantages. The first one is the aforementioned parsimony that in some cases is a necessary condition to estimate a model. The second one is that the factor structure in the regressor coefficient permits to link the dependent and independent variables through a restricted number of multilinear relationships. Indeed, this specification can be seen as a reduced rank regression with multilinear interactions. This is very desirable feature when we deal with rank deficient (or near rank deficient) tensors or when we want to forecast, because the links between regressor and response are restricted only to fewer strong (multilinear) relations. In addition, the Tucker structure, contrarily of the PARAFAC, can handle dimension asymmetric (modality skewness) tensor regressions since each dimension do not need to have the same number of components.
	\subsection{Estimation}
	In this section we introduce the supervised learning procedure to estimate the parameters of model presented in Eq. \ref{tr}. Following the standard literature in regression analysis, we select the model parameters which minimise the sum of squared residuals\footnote{Without loss of generality we omit $\T{A}$ from the derivation.}
	\begin{equation}\label{opt}
		\|\T{Y}- \langle \T{X},\T{B} \rangle_{(\T{I_x};\T{I_B})}\|^2_F=\|\T{E}\|^2_F=\langle \T{E},\T{E} \rangle
	\end{equation}  
	Imposing the Tucker structure of Eq. \ref{tb} in the coefficient the optimization takes the form
	
	\begin{equation}\label{optimization}
\T{\widehat{B}}=\argmin_{Trk(\T{B})\leq \T{R_{\bullet}}}\|\T{Y}- \langle \T{X},\T{B} \rangle_{(\T{I_x};\T{I_B})}\|^2_F
	\end{equation}
	where $Trk(\T{B})$ is the Tucker rank of $\T{B}$ and $\T{R_{\bullet}}$ is the chosen/estimated dimension of the core tensor $\T{G}$. The learning of the model parameter is a nonlinear optimization problem which cannot be solved all at once. However, it can be solved by iterative algorithms such as Alternating Least Squares (ALS). The ALS algorithm was introduced for the Tucker decomposition by \cite{kroonenberg1980principal,kapteyn1986approach}. This methodology solves the full optimization problem by dividing it into small least squares problems. This allows to solve the optimization problem described in Eq. \ref{optimization} for each factor matrix (and the core tensor $\T{G}$) of the Tucker structure separately, keeping constant the other ones at the previous iteration value. Then, it iterates alternating among all the Tucker components until the algorithm converges (a minimum error threshold is reached) or a maximum number of iterations is exceeded. 	The partial least square solutions over each component $\Mn{U}{n}$ and $\T{G}$ for the general case of an $N$-th order coefficient tensor are presented in Algorithm \ref{ALS1}. Iterating over each component taking the others fixed, permits to solve the \textit{\textquotedblleft big\textquotedblright} problem by a set of \textit{\textquotedblleft small\textquotedblright} least squares problems.
	
	\begin{algorithm}[H]
		\caption{Alternating least squares}\label{ALS1}
		\begin{algorithmic}[1]
			\State \textbf{Initialize the algorithm to some $\Mn{{U}}{n}_{0}$ and $\T{G}_0$.}
			\Repeat
			\State $\Mn{{U}}{1}_{i}=\phi(\Mn{{U}}{2}_{i-1},\dots,\Mn{{U}}{N}_{i-1},\Mn{{V}}{1}_{i-1},\dots,\Mn{{V}}{M}_{i-1},\T{G}_{i-1},\T{X},\T{Y})$

				\State\vdots
				
			\State$\Mn{{U}}{N}_{i}=\phi(\Mn{{U}}{1}_{i},\dots,\Mn{{U}}{N-1}_{i},\Mn{{V}}{1}_{i-1},\dots,\Mn{{V}}{M}_{i-1},\T{G}_{i-1},\T{X},\T{Y})$

			\State$\Mn{{V}}{1}_{i}=\phi(\Mn{{U}}{1}_{i},\dots,\Mn{{U}}{N}_{i},\Mn{{V}}{2}_{i-1},\dots,\Mn{{V}}{M}_{i-1},\T{G}_{i-1},\T{X},\T{Y})$

			\State\vdots

			\State	
		$\Mn{{V}}{M}_{i}=\phi(\Mn{{U}}{1}_{i},\dots,\Mn{{U}}{N}_{i},\Mn{{V}}{1}_{i},\dots,\Mn{{V}}{M-1}_{i},\T{G}_{i},\T{X},\T{Y})$

			\vspace{5pt}

			\State$\T{G}_{i}=\phi(\Mn{{U}}{1}_{i},\dots,\Mn{{U}}{N}_{i},\Mn{{V}}{1}_{i},\dots,\Mn{{V}}{M-1}_{i},\T{X},\T{Y})$
			
			\Until {Convergence or Maximum iterations reached.}
			\State \textbf{Return $\widehat{\T{B}}$}
		\end{algorithmic}
	\end{algorithm}
	
	 In  Algorithm \ref{ALS1}, $\phi(\cdot )$ is the LS problem for each subcomponent and it reveals the dependence of each sub-problem to the remaining components. We now explicitly show the LS solution for each Tucker component, i.e. $\Mn{{U}}{1}$,$\dots$, $\Mn{{U}}{N}$, $\Mn{{V}}{1}$,$\dots$, $\Mn{{V}}{M}$ and $\T{G}$. We start by estimating the input modes matrices $\Mn{{U}}{1}$,$\dots$,$\Mn{{U}}{N}$. We show the solution of the general $\Mn{{U}}{n}$ taking the other components fixed. Define $\T{H}_{[n]}$ to be the contracted product between $\T{X}$ and the Tucker coefficient $\T{B}$ without the $n$-th component matrix, i.e.

\begin{subequations}\label{eqB}
	\begin{align}
	  \T{H}_{[n]} & = \langle\T{X},\T{B}_{-n} \rangle_{(\T{I}_{\T{X}_{n+1}};\T{I}_{\T{B}_n})}         \label{eqB1} \\
	 \begin{split}
		\T{B}_{-n} & = \T{G} \times_1 \Mn{U}{1} \cdots \times_{n-1} \Mn{U}{n-1} \times_{n+1} \Mn{U}{n+1} \cdots \times_N \Mn{U}{N} \cdots\\
	& \times_{N+1} \Mn{V}{1} \cdots \times_{N+M} \Mn{V}{M} 
	\end{split}
	\label{eqB2}
	\end{align}
\end{subequations}
	where $\T{I}_{\T{X}_{n+1}}$ and $\T{I}_{\T{B}_n}$ are the set of indices without the mode corresponding to the $n$-th component matrix. The resulting tensor $\T{H}_{[n]}$ is of dimension $N \times I_n \times F_n \times J_1 \times \cdots \times J_M $. Unfolding  $\T{H}_{[n]}$ over the $2$-nd and $3$-rd modes, we get $ \M{H}_{[n]} \in \mathbb{R}^{I_n F_n\times N \T{M} }$ where $\T{M}=\prod_{i=1}^M J_i$. This is the design matrix used to predict $\text{vec}(\T{Y})$ from all the columns of $\Mn{{U}}{n}$.\footnote{ One can alternatively compute the $\M{H}^{i}_{[n]}$ matrix related to the $f^{i}_n$-th ($f^{i}_n=1,2,\dots, F_n $) column of $\Mn{{U}}{n}$ and then concatenate all these matrices to get $ \M{H}_{[n]}=[ \M{H}^{1}_{[n]},\dots,\M{H}^{F_n}_{[n]}]$.} Consequently, the LS solution for the $n$-th input mode matrix is
	\begin{equation}
	\text{vec}(\Mn{{U}}{n})=(\M{H}_{[n]} \M{H}_{[n]}^T)^{-1} \M{H}_{[n]} \text{vec}(\T{Y}).
	\end{equation}
	
	The procedure to find the LS solution of the outcome modes is similar. We show the result for the general $\Mn{V}{m}$ matrix taking all the other Tucker components fixed. Define $\T{O} _{[m]}$ to be the contracted product between $\T{X}$ and the tucker coefficient $\T{B}$ without the $m$-th outcome component matrix, i.e.
	
	\begin{subequations}\label{eqmB}
		\begin{align}
		 \T{O}_{[m]} & = \langle\T{X},\T{B}_{-m} \rangle_{(\T{I}_{\T{X}};\T{I}_{\T{B}})}         \label{eqmB1} \\
		 \begin{split}
		 \T{B}_{-m} & =  \T{G} \times_1 \Mn{U}{1}  \cdots \times_N \Mn{U}{N} \times_{N+1} \Mn{V}{1} \cdots  \times_{m-1} \Mn{V}{N+m-1} \cdots\\
		& \times_{N+m+1}  \Mn{V}{m+1} \times_{N+M} \Mn{V}{M} 
		\end{split}
		\label{eqmB2}
		\end{align}
	\end{subequations}
	where $\T{I}_{\T{X}}$ and $\T{I}_{\T{B}}$ are the full set of indices already defined for Eq. \ref{tr}. The resulting tensor $\T{O}_{[m]}$ is of dimension $N \times J_1 \times \cdots \times J_{m-1} \times F_m \times J_{m+1} \times J_M $. Unfolding  $\T{O}_{[m]}$ over the $(m+1)$-th mode gives $ \M{O}_{[m]} \in \mathbb{R}^{F_m \times N \T{M}_m }$ where \[\T{M}_m=\prod_{\shortstack{$\scriptstyle i=1 $\\$\scriptstyle i\neq m$}}^M J_i.\] This is the design matrix used to predict the unfolded response tensor $\T{Y}$ along the $(m+1)$-th outcome mode, i.e. $\M{Y_{(m+1)}} \in \mathbb{R}^{J_m \times N\T{M}_m }$. Therefore, the LS solution for the $m$-th outcome mode matrix is
	\begin{equation}
	\Mn{{V}}{m}=(\M{O}_{[m]} \M{O}_{[m]}^T)^{-1} \M{O}_{[m]} \M{Y^T_{(m+1)}}.
	\end{equation}
	Finally, the unfolded core tensor is updated as:
	\begin{equation}
	\M{G}=(\M{X^\star}_{(1)}^T \M{X^\star}_{(1)})^{-1} \M{X^\star}_{(1)}^T \M{Y^\star}_{(1)}.
	\end{equation}
	Where $\M{X^\star}_{(1)}$ and $\M{Y^\star}_{(1)}$ are the $1$-st mode matricization of
	
	\begin{subequations}\label{eqmG}
		\begin{align}	
	&	\T{X}^{\star}= \T{X} \times_{2}  \Mn{{U}}{1}  \cdots \times_{N+1}  \Mn{{U}}{N} \label{eqnG1} \\		
		&	\T{Y}^{\star}= \T{Y} \times_{2} \Mn{{V}}{1}_{ \dagger}   \cdots \times_{M+1}  \Mn{{V}}{M}_{\dagger} \label{eqmG1}		
		\end{align}	
	\end{subequations}

	where $ \M{V}_{\dagger}=\M{V}^{\dagger}$ and $\dagger$ is the Moore–Penrose pseudo-inverse. If the dataset is not centered, the intercept tensor $\T{A}$ is computed as:
	
	\[\T{\widehat{A}}= \mathbb{E}[\T{Y}] - \langle \mathbb{E}[\T{X}], \T{\widehat{B}}  \rangle_{(\T{I_x};\T{I_B})}. \]

	\subsection{Penalized Tensor regression}
	Even if the reduced rank specification is able to reduce the redundant information across the tensor dimensions, it is still prone to over-fitting when intra-mode collinearity is present. In this case, a shirkage estimator is necessary for a stable solution. Indeed, the presence of collinearity between the variables of the dataset degrades the forecasting capabilities of the regression model. In this work we will use the Tikhonov regularization \cite{tikhonov1943stability,tikhonovsolutions,kennedy2003guide}. Known also as Ridge regularization, it modifies the optimization problem in Eq. \ref{opt} as	
		\begin{equation}\label{optm2}
	\T{\widehat{B}}=\argmin_{Trk(\T{B})\leq \T{R_{\bullet}}}\|\T{Y}- \langle \T{X},\T{B} \rangle_{(\T{I_x};\T{I_B})}\|^2_F + \lambda	\|\T{B}\|^2_F,
	\end{equation}
where $\lambda>0$ is the regularization parameter and $\| \|^2_F$ is the squared Frobenius norm. The greater the $\lambda$ the stronger is the shrinkage effect on the parameters. However, high values of $\lambda$ increase the bias of the tensor coefficient $\T{B}$. For this reason, the shrinkage parameter is usual set via data driven procedures rather then input by the user. The Tikhonov regularization can be computationally very expensive for big data problem. To solve this issue, \cite{arcucci2017decomposition} proposed a decomposition of the Tikhonov regularization that showed to be computationally very effective. The optimization problem presented in Eq. \ref{optm2} can be rewritten as an unregularized problem with modified regressor and response tensors. In particular, we can rewrite Eq. \ref{optm2} as

	\begin{equation}\label{optm3}
\T{\widehat{B}}=\argmin_{Trk(\T{B})\leq \T{R_{\bullet}}}\|\T{\bar{Y}}- \langle \T{\bar{X}},\T{B} \rangle_{(\T{I_x};\T{I_B})}\|^2_F, 
\end{equation}
where $\T{\bar{X}}$ is the modified version of $\T{X}$. Using the same argument in \cite{lock2017}, we can exploit multilinear algebra and easily rewrite the LS solution without the need of running the regression with the augmented predictor and response variables, which can be high dimensional. The solution for the general $\Mn{{U}}{n}$ taking the other components fixed is given by

	\begin{equation}
	\text{vec}(\Mn{{U}}{n})=(\M{H}_{[n]} \M{H}_{[n]}^T+  \M{I}_{I_n \times I_n} \otimes \lambda \langle \T{B}_{-n}, \T{B}_{-n}\rangle_{(n;n)}  )^{-1} \M{H}_{[n]} \text{vec}(\T{Y})
	\end{equation}
where $\M{I}_{I_n \times I_n}$ is the identity matrix of dimension $I_n \times I_n$. For the outcome modes matrices the solution becomes

\begin{equation}
\Mn{{V}}{m}=(\M{O}_{[m]} \M{O}_{[m]}^T+\lambda \langle \T{B}_{-m}, \T{B}_{-m}\rangle_{(m;m) })^{-1} \M{O}_{[m]} \M{Y^T_{(m+1)}}.
\end{equation}

The contracted products $\langle \T{B}_{-n} \T{B}_{-n}\rangle_{(n;n)}$ and $\langle \T{B}_{-m} \T{B}_{-m}\rangle_{(m;m)}$ result in matrices of dimensions $F_n \times F_n$ and $F_m \times F_m $ respectively. In this work we leave the $\T{G}$ unregularized since we want that the core tensor, which regulates the interconnections between the tensor dimensions, not to be biased. However, the Ridge solution of $\M{G}$ takes the form 

	\begin{equation}
\M{G}=(\M{X^\star}_{(1)}^T \M{X^\star}_{(1)}+\lambda \M{I})^{-1} \M{X^\star}_{(1)}^T \M{Y^\star}_{(1)}.
\end{equation} 	
	\section{Simulation study}\label{sec_ss}
	
\begin{figure}[h!]
	\begin{center}	
			\includegraphics[height=0.45\textwidth, width=0.5\textwidth]{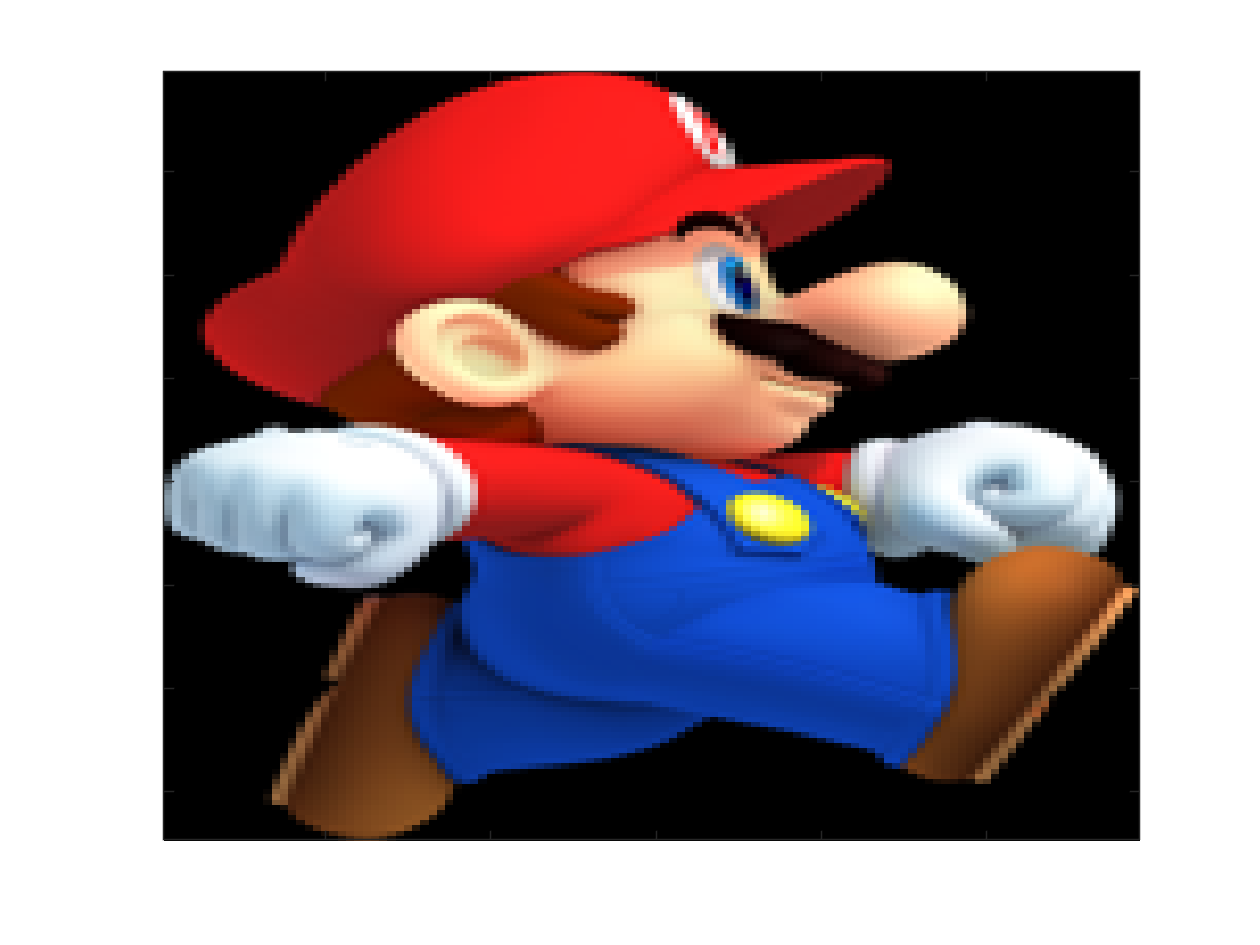}
			\vspace{-15pt}
			\caption{Image coefficient tensor $\T{B}$.}
			\label{fig:fig1}		
\end{center}
\end{figure}

	In this section, we test the reliability of the model and its estimation procedure through a simulation study. For this purpose, we run two experiments. The first one dedicated to the ability of the model to retrieve the real, structural $\T{B}$ coefficient with a compressed Tucker version. To do this, we use as structural $\T{B}$ the image presented in Fig. \ref{fig:fig1} which is a $3$-rd order tensor of dimension $149 \times 118 \times 3$. The first two dimensions relates to the activation of the image's pixels while the third mode controls the colour of each pixel. 

\begin{center}
	 	\begin{figure}[h!]
	 			 		\includegraphics[height=0.85\textwidth, width=1\textwidth]{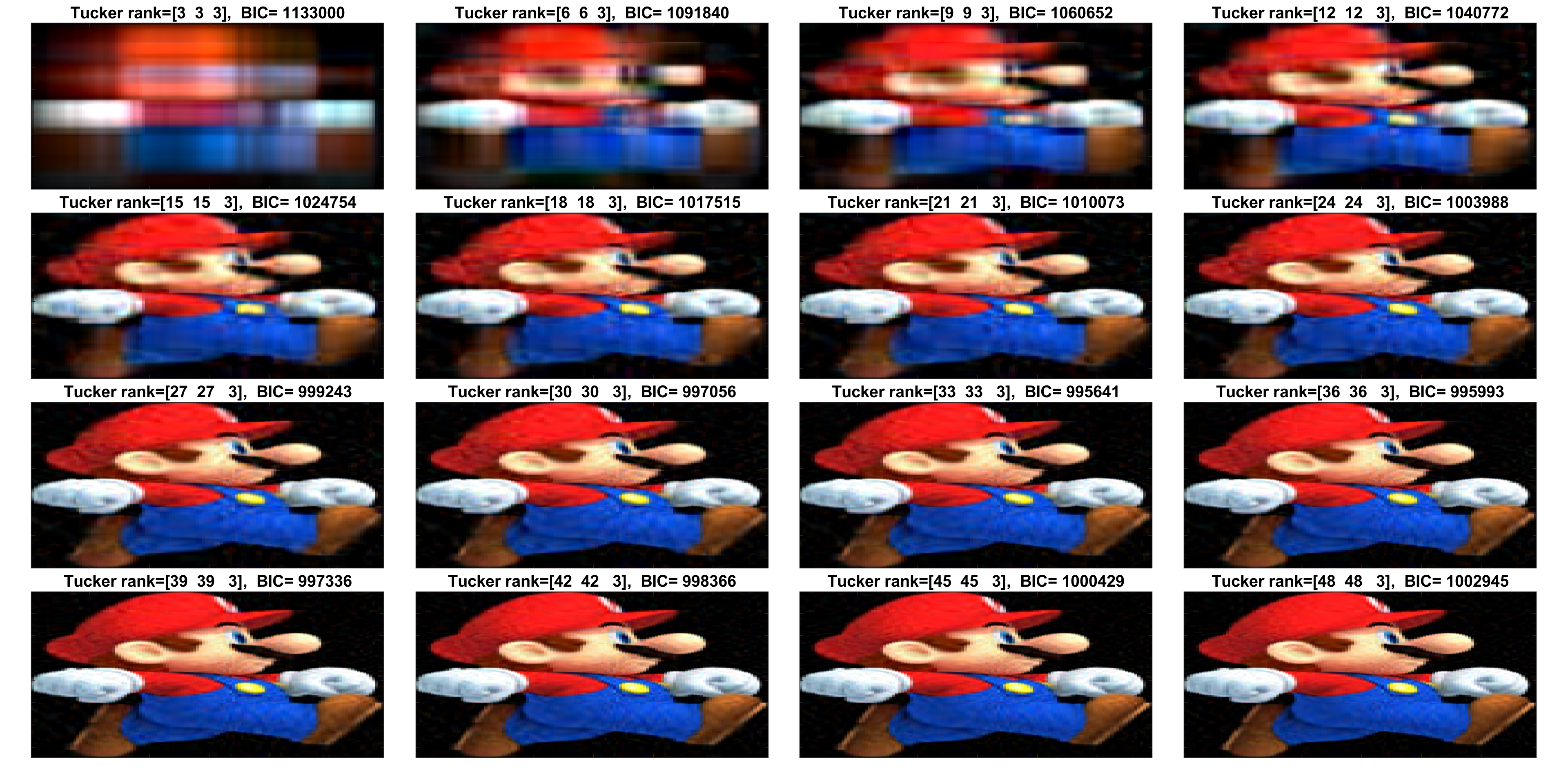}
	 		\caption{Results for the tensor regression with image coefficient tensor $\T{B}$. Numbers between square brackets refer to the size of the core tensor. The data-driven BIC criterion of Eq. \ref{eq_BIC} are also reported.}
	 		\label{fig:fig2}
	 	\end{figure}	 	
	 \end{center}

	We then generate a regression dataset as follows:
	\begin{itemize}
	\item[-] Generate $\T{X} \in \mathbb{R}^{250 \times 149}$ from $N(0,1)$.
	\item[-] Generate $\T{E} \in \mathbb{R}^{250 \times 118\times 3}$ from $N(0,1)$.
\item[-] $\T{Y}=\langle \T{X},\T{B} \rangle_{(2;1)}+\T{E}$.
	\end{itemize}
	The resulting $\T{Y} $ is of dimension $250 \times 118\times 3$. In Fig. \ref{fig:fig2} we report the tensor regression coefficient $\T{B}$ corresponding to different core specifications and their respective Bayesian information criterion (BIC). The Bayesian information criterion  \cite{schwarz1978} is a data-driven method for model selection. It puts into relationship the goodness of fit of a model and its complexity, penalizing non-parsimonious models. It is computed as follows:
	
	\begin{equation}\label{eq_BIC}
	\text{BIC}= u\ln\left( \frac{SSR}{u}\right) +w\ln(u)	
	\end{equation}	
	 where $u$ is the number of data-points, $w$ is the number of estimated elements in the Tucker coefficient $\T{B}$ and $SSR$ is the sum of squared residuals of the regression. The BIC is formed by two additive components. The first one tends to decrease as the model becomes more complex, since it will reduce the sum of squared residuals while the second component increases as the model complexity increases. Therefore, the best model is associated with the minimum BIC. As can we see from Fig. \ref{fig:fig2}, the BIC is minimised for the setting corresponding to a core tensor of size $33 \times 33 \times 3$. This is a considerably high compression with respect to the original dimension ($\approx$77\% compression). Moreover, from a graphical point of view, the reconstruction seems pretty accurate.   
	For the second experiment, we build a tensor dataset with strong collinearity within the modes. Collinearity is a condition for which different regressors are strongly correlated, making the estimation unstable. To simulate the dataset we use the Array Normal model with separable covariance matrices introduced in \cite{hoff2011separable,akdemir2011array}. The model uses the Tucker product to generate a tensor for which each dimension has its own covariance matrix. Take a tensor  $\T{X}\in \mathbb{R}^{T \times  I_1 \times \cdots \times I_N}$.  $\T{X}$ is an Array Normal dataset with covariance matrices $\Sigma_{T}$, $\Sigma_{I_1}$, $\dots$, $\Sigma_{I_N}$ if we can rewrite it as:	
	\begin{equation}\label{anormal}
	\T{X}=\T{Z}\times_1 \Sigma^{\frac{1}{2}}_{T} \times_2  \Sigma^{\frac{1}{2}}_{1}    \cdots \times_{N+1} \Sigma^{\frac{1}{2}}_{N}
	\end{equation} 
	
	where $\T{Z}$ is a normal variate tensor and $\Sigma^{\frac{1}{2}}_{i}$ is the square root of the covariance matrix for the $i$-th dimension. For this simulation, we  generate a regression dataset as follows:
	\begin{itemize}
		\item[-] Generate a standard Gaussian tensor $\T{Z} \in \mathbb{R}^{100 \times 6 \times 19}$.
		\item[-] Generate the Covariance matrix for mode $i$ ($i=T,I_1,I_2$) as $\Sigma_{i}=\rho_i^{|p-q|}$, where $p=1,\dots,P$ and $q=1,\dots,Q$ are the indices of the matrix components.
		\item[-] Compute the  $\Sigma^{\frac{1}{2}}_{i}$ as the Cholesky decomposition of $\Sigma_{i}$.		
		\item[-] Build $\T{X}$ as in Eq. \ref{anormal}.
		\item[-] Generate $\T{B}$ to be Tucker tensor of dimension $ 6\times 19 \times 6\times 19$ with core tensor of size $2\times 3 \times 2\times 3$.
		\item[-] Generate $\T{E} \in \mathbb{R}^{100 \times 6\times 19}$ from $N(0,1)$.
	\end{itemize}
	
	The response tensor $\T{Y}$ is then formed as
		\[\T{Y}=\langle \T{X},\kappa\T{B} \rangle_{(2,3;1,2)}+\T{E}\]
	where $\kappa$ is a scalar such that the Signal to Noise Ratio (SNR) is equal to a specified value, i.e. \[\text{SNR}=\frac{\|\langle \T{X},\kappa\T{B} \rangle_{(2,3;1,2)}\|_F^2}{\|\T{E}\|_F^2}.\] The resulting $\T{Y}$ is of dimension $100 \times 19\times 6$. For the covariance matrices we use the following value of $\rho_T=0.1$, $\rho_1=0.95$ and $\rho_2=0.8$ assuming strong collinearity in the second mode and a moderate one in the third mode. To see the effect of the shrinkage on prediction, we generate a new dataset of the same dimension and with the same $\T{B}$, i.e.
		\[\T{Y}_{\text{new}}=\langle \T{X}_{\text{new}},\kappa\T{B} \rangle_{(2,3;1,2)}+\T{E}_{\text{new}}.\]
	We then compute the BIC for different specifications of the model presented in Eq. \ref{optm2}, using $\T{Y}_{\text{new}}$ and $\T{\widehat{Y}}_{\text{new}}=\langle \T{X}_{\text{new}},\T{\widehat{B}} \rangle_{(2,3;1,2)}$ for the computation of the SSR. For this simulation study we used the fowling specifications:
	\begin{itemize}
\item[-] Tucker rank $\T{R_{\bullet}}=[f, g, f, g]$ with $f=1,\dots,6$ and $g=1,\dots,19$.
\item[-] Shrinkage parameter $\lambda=0,0.5,1,5,50$ as in \cite{lock2017}.
\item[-] $\text{SNR}=1,5$ as in \cite{lock2017}.
	\end{itemize}
	Table \ref{tab1} provides the results for the simulation experiment. Given the large dimension of the result, we provide for each $\lambda$ and SNR, the minimum value of the BIC and the corresponding estimated $\T{R_{\bullet}}$.
	
	\begin{table}[h!]
			\begin{center}
		\begin{tabular}{|l|c|c|c||c|c|c|}
			\hline
			\multirow{2}{*}{} & \multicolumn{3}{c||}{SNR=1}                                                                 & \multicolumn{3}{c|}{SNR=5}  \\ \cline{2-7} 
			& \multicolumn{1}{c|}{BIC} & \multicolumn{1}{c|}{$\hat{f}$} & \multicolumn{1}{c||}{$\hat{g}$} & BIC & $\hat{f}$ & $\hat{g}$ \\ \hhline{|=|=|=|=|=|=|=|}
			$\lambda=0$       &        1.7924e+03  &        2                        &        3                        &     1.9677e+04  &    1       &        1   \\ \hline
			$\lambda=0.5$     &       \textbf{1.7792e+03}   &         2                       &         3                       &    1.9630e+04  &     1      &    1       \\ \hline
			$\lambda=1$       &       1.7847e+03  &           2                     &           3                     &     1.9595e+04 &    1       &      1     \\ \hline
			$\lambda=5$       &       1.9243e+03  &            2                    &            3                    &  1.9438e+04   &   1        &      1     \\ \hline
			$\lambda=50$      &        2.4547e+03  &            2                    &            3                    &   \textbf{1.8134e+04 }  &  1         &     2      \\ \hline
		\end{tabular}
	\vspace{3pt}
		\caption{Table for the BIC computed on predicted values. }
	\label{tab1}
\vspace{-8pt}	
\end{center}
	\end{table}
As it is possible to notice, for the low SNR scenario, the model has its optimum in correspondence to a shrinkage parameter $\lambda=0.5$ and $\T{\widehat{R}_{\bullet}}=[2, 3, 2, 3]$ which is the core tensor size we used to create the regression. In the case in which the $SNR=5$, the model strongly shrinks the coefficient tensor ($\lambda=50$) and the low-rank approximation suggests a one-factor model in most of the cases. This could be due to the fact that the redundant information is very strong due to the importance of the numerator of the SNR has with respect to the denominator and the model captures this phenomenon. With these results we showed in two simulation experiments that the model is able to correctly estimate the true structural coefficient even with a strong compression and to predict the response tensor variable in presence of redundant information induced by the collinearity condition. In the next section, we apply the tensor regression model on real data.
	\section{Empirical application}\label{sec_ea}
	In this section, we apply the proposed tensor regression model described in Section \ref{sec_tr} to real data in a forecasting application. The purpose of this analysis is to test the prediction performance of the model against existing ones in different contexts. To this aim, we carry out two applications with the use of two different datasets. The first one is a spatio-temporal dataset based on the Foursquare platform. This dataset has been already used in the machine learning literature to test the forecasting ability of multilinear multitasking learning models \cite{yu2015}. For this application, we compare the  Accelerated Low-rank Tensor Online Learning (ALTO) model proposed in \cite{yu2015} with our tensor regression model presented in this paper. The second application consists in forecasting a panel composed by 6 macroeconomic time series for 19 countries. In this case, we use as benchmark model the Vector Autoregression (VAR), which has become the workhorse model for macroeconomic forecasting \cite{var_book}. In both cases, the tensor regression model becomes a tensor autoregressive model as the predictor tensor is a lagged version of the response tensor. 
	\subsection{Datasets}
	In this subsection, we provide details of the dataset used in the empirical applications.
	\subparagraph{Foursquare dataset:}
	Foursquare is one of the most popular recommendation systems in the world where users can query venues such as hotels, universities and restaurants. Additionally, Foursquare users can share their check-ins and reviews with other users, which make it different from standard search engines. On Foursquare, a person can check-in at his current location, leave tips about the venue, explore discounts around his current location, or add other people as their friends. For these reasons, the dataset has been used in different research contexts, as for example for recommendation systems and urban mobility analysis. The Foursquare dataset we use contains the users' check-in records in the Pittsburgh area between 24/02/2012 and 24/05/2012. The dataset is composed by the number of check-ins by 121 users belonging to 15 different types of venues over 1198 time periods, resulting in a $3$-rd order tensor of dimension $121 \times 15 \times 1198$. 
	\subparagraph{Macroeconomic panel dataset:}
	The second dataset used is a subsample of the data presented in \cite{pesaran2009} on Global VAR. The subsample contains quarterly data from 1979Q2 to 2016Q4 of real GDP growth, the rate of inflation, log-returns of the nominal equity price index, log-returns of the exchange rate, nominal short-term interest rate and nominal long-term interest rate. Countries in the subsample (the ones for which all data is available) are: Australia, Austria, Belgium, Canada, France, Germany, India, Italy, Japan, Korea, Netherlands, Norway, New Zealand, South Africa, Spain, Sweden, Switzerland, UK, USA. A fully detailed review of the full dataset has been published in \cite{mohaddes2018compilation}. Let us stress that the use of multicountry data is useful in capturing inter-countries' interrelations rather than only intra-countries' interactions. This is particularly useful in forecasting exercises since some countries' variables are strongly influenced not only by their past but also from other countries' economic indicators. Along these lines, the proposed model exploits this information in a reduced rank prototype model, where the countries' modes are interrelated with the variable modes through the Tucker specification. We have checked the data for stationarity via the KPSS test and the Augmented Dickey Fuller test. All time series results to be trend stationary at $5\%$ confidence level. The resulting dataset is a tensor of dimension $150 \times 6 \times 19$ variables hence the $\T{B}$ tensor coefficient to be estimated in the autoregressive model is of dimension $6 \times 19 \times 6 \times 19$ that implies $12996$ parameters (assuming no intercept).
	
	\subsection{Forecasting}
	Let us consider the tensor regression where the predictor ($\T{X}$) is a lagged version of the response variable ($\T{Y}$). This is the case of Autoregression in which a set of variables are regressed on their past values. In this respect, the tensor regression takes the form of a Tensor Autoregression (TAR). In the case of a Tensor Autoregressive model, the coefficient tensor can be seen as a sort of Multilayer causality network in which each coefficient determines the effect in time and space from one variable to itself and to the other variables \cite{brandi2020new}. To test the TAR forecasting ability and compare it with alternative models (ALTO and VAR), we compute the mean root square forecasted error for each time series across different models, i.e.:
	
	\[\text{RMSFE}_h=\sqrt{ \frac{\sum_{h=1}^{H}(\widehat{Y}_{t+h}-Y_{t+h})^2}{h}}\]
	
	where $h$ is the forecasting horizon. The model with a lower $\text{RMSFE}_t$ is the one with a better performance. 
	
	\subsection{Empirical results}
	In this section, we show the empirical results of the Tensor Autoregression in the two applications. 
	\subsubsection{Foursquare dataset}
	For the Foursquare dataset, we compare the RMSFE of the Accelerated Low-Rank Tensor Learning (ALTO) model proposed in \cite{yu2015}\footnote{The data and the Matlab code are available at the webpage of Rose Yu, http://roseyu.com/code.html} with the TAR model. We use the authors' code for the ALTO model and compare the RMSFE of the two models with up to 4 lags and Tucker rank up to 4. Table \ref{fs} summarizes the results obtained using a 80\% of training set.\footnote{The training set is the subsample corresponding to the segment of the data that is used to estimate the model coefficients. The remaining part, the test set, is used to compute the forecasting error.} 

	\begin{table}[h!]
		\begin{center}
		
		\begin{tabular}{|l|l|l|l||l|l|l|}
			\hline
			& \multicolumn{3}{c||}{ALTO} & \multicolumn{3}{c|}{TAR} \\ \hline
			\diagbox[width=1.5cm, height=1.cm]{\scriptsize{\textbf{Lag}}}{ \hspace*{\fill}\scriptsize{\textbf{ Rank}}}& \multicolumn{1}{c|}{1}       & \multicolumn{1}{c|}{2}      & \multicolumn{1}{c||}{3}      & \multicolumn{1}{c|}{1}      & \multicolumn{1}{c|}{2}      & \multicolumn{1}{c|}{3}      \\ \hline
			1 &  0.13737    &   0.16135    &   0.16556    &    0.12689    &    0.12398   &   0.12353     \\ \hline
			2 &  0.12890    &   0.14400   &   0.14641     &   0.12551     &    0.12400    &     0.12364   \\ \hline
			3 &  0.12997    &   0.13887   &   0.14001     &   0.12454     &   0.12418     &    0.12380    \\ \hline
			4 &  0.12892    &   0.13547   &   0.13572   &     0.12471   &    0.12420    &    0.12440    \\ \hline
		\end{tabular}
	\caption{RMSFE of the ALTO and TAR models for the Foursquare data with different lag and Tucker rank specifications.}\label{fs}
	\vspace{-10pt}
		\end{center}
	\end{table}

The result suggests that the TAR model outperforms the ALTO at any lag specification and in all tensor rank specifications considered. This can be attributed to two main reasons. The first one is the fact that the ALTO is a 2-steps procedure in which in the first step the model is run as 15 independent VAR models (one for each venue) and then, after concatenating the coefficient matrices, a tensor low-rank constraint is imposed. This modelling procedure does not exploit the full potential of the reduced rank regression since the low-rank constraint is imposed only after a full rank model is estimated. The second issue is that the 2-steps procedure does not take into account the interconnections between the 15 venues since those are run independently. The post-estimation low-rank constraint tackles only marginally the problem. In contrast, the Tensor Autoregressive model exploits all interdimensional information estimating a low-rank model all at once. 
	
	\subsubsection{Macroeconomic panel data}
\begin{figure}[ht!]

		\begin{subfigure}[t]{0.49\textwidth}
					\hspace*{-1.5cm}
		\includegraphics[height=0.95\textwidth, width=1.25\textwidth]{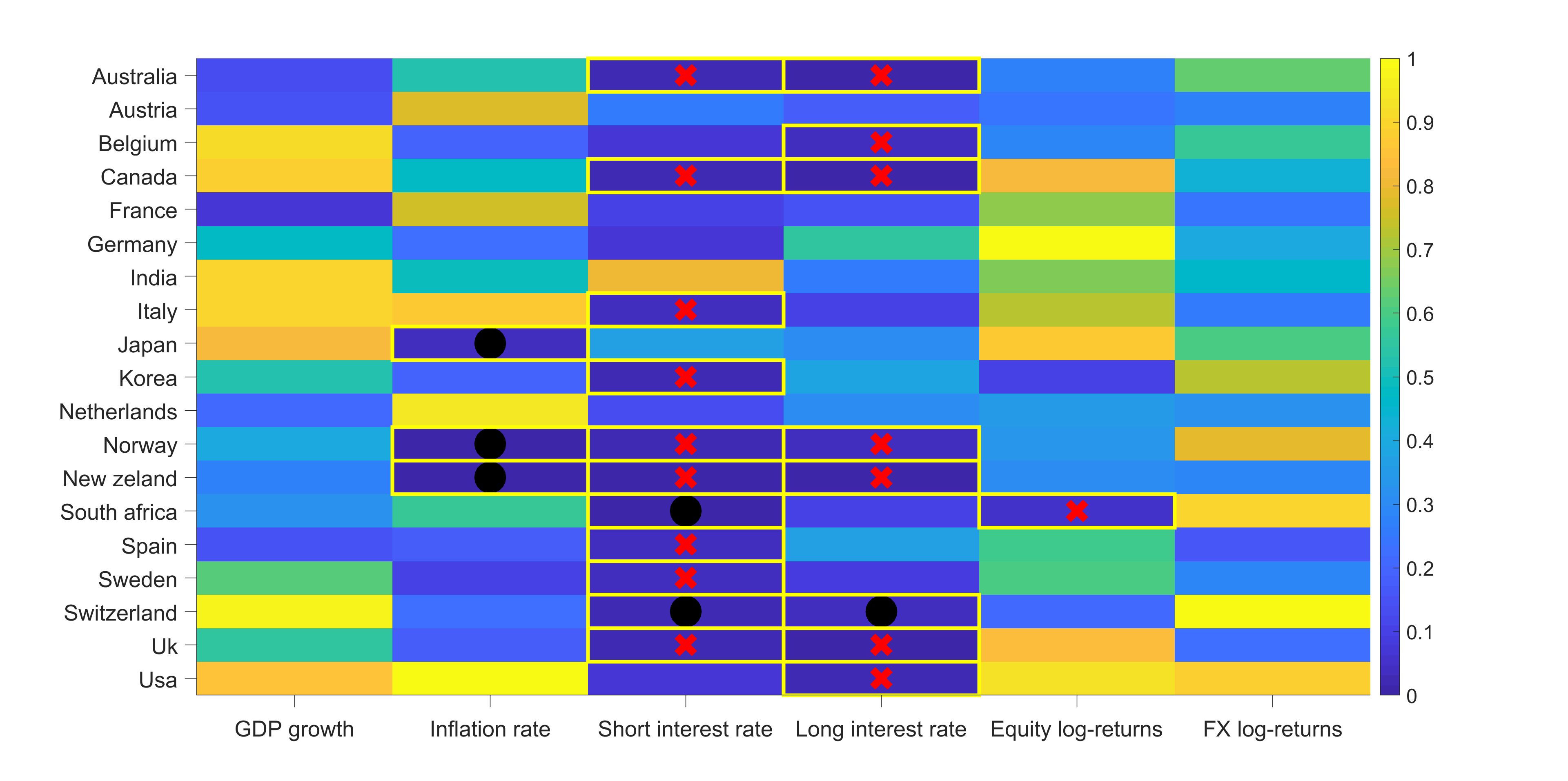}
		\vspace{-25pt}
		\caption{1 step ahead forecast}
		\end{subfigure}
	\hspace{-15pt}
	\begin{subfigure}[t]{0.49\textwidth}
		\includegraphics[height=0.95\textwidth, width=1.25\textwidth]{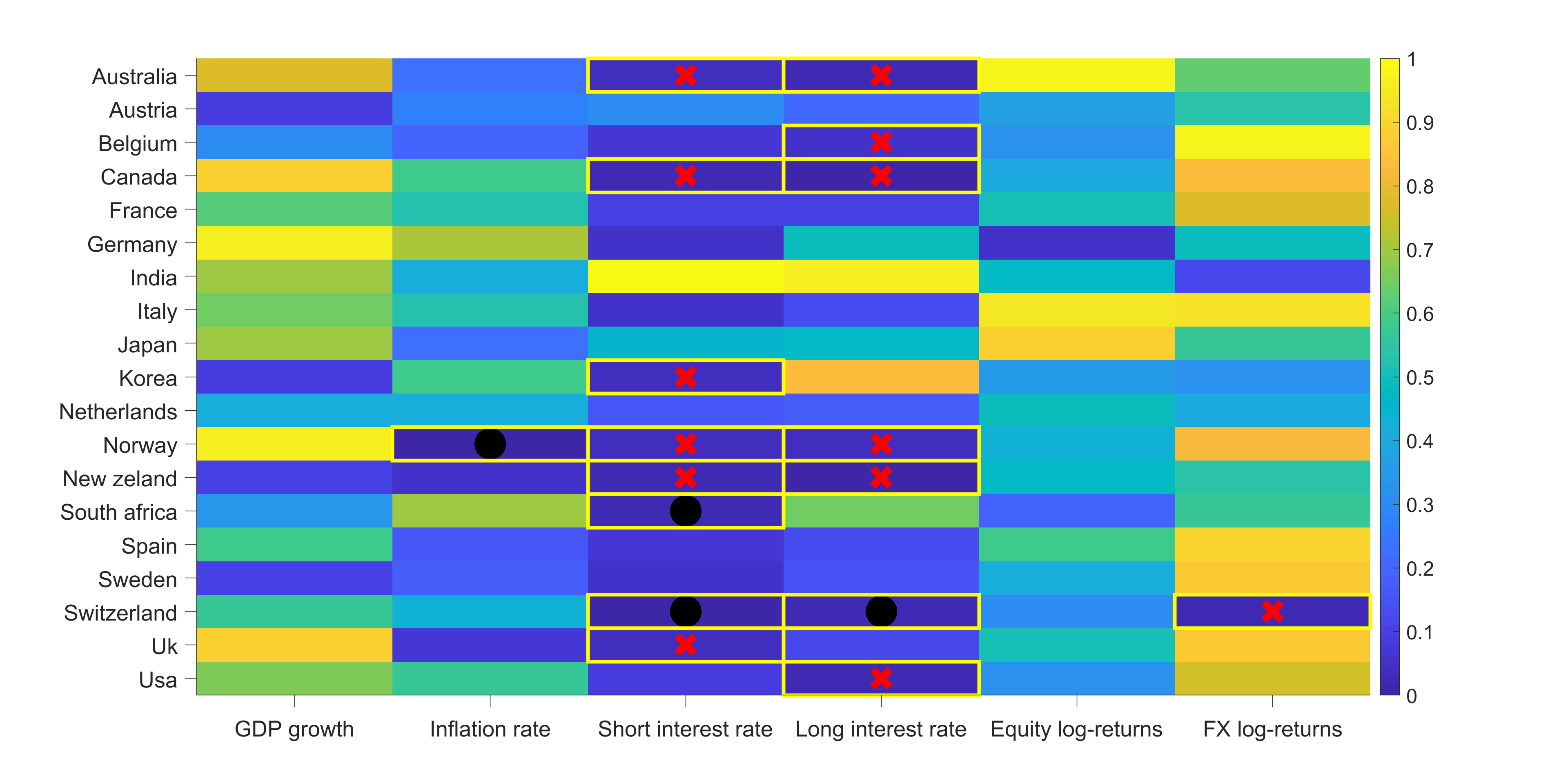}				
					\vspace{-25pt}
						\caption{2 steps ahead forecast}
						
						\hspace*{-1.5cm}
	\end{subfigure}
		\vspace{4pt}
		\caption{ P-values of the DM test between TAR\{1; [1, 1, 1, 1]\} and the VAR(1). Yellow rectangles denote p-values lower than 0.05. Red crosses (\textcolor{red}{$\mathbf{\times}$}) refer to cases in favour of the TAR model while black circles  (\tikzcircle{3pt}) refer to cases in favour of the VAR.}
		\label{fig:fig4}		

\vspace{5pt}
\end{figure}

\begin{figure}[ht!]

	\begin{subfigure}[t]{0.49\textwidth}
		\centering
		\hspace*{-1.5cm}
		\includegraphics[height=0.95\textwidth, width=1.25\textwidth]{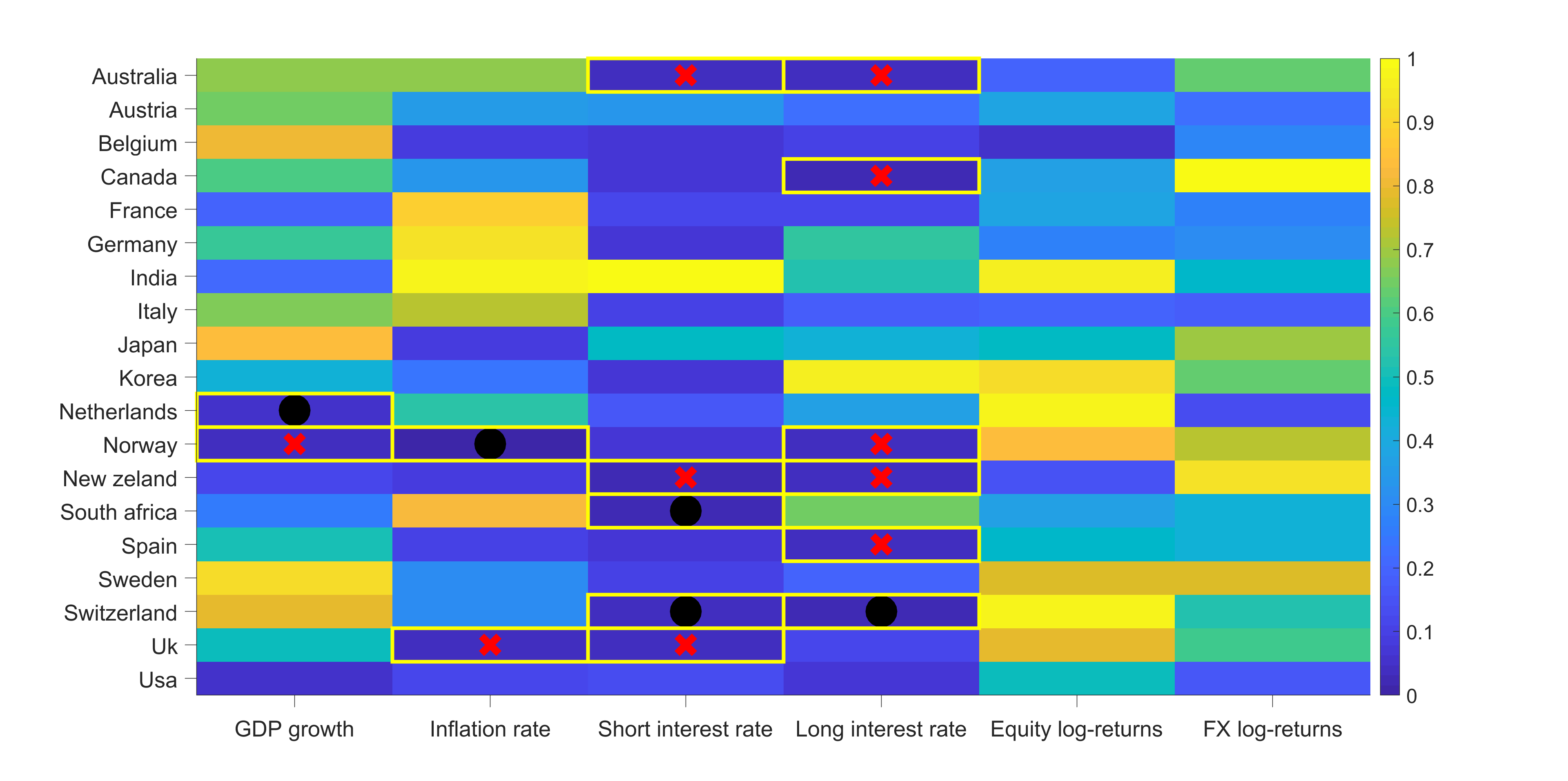}
		\vspace{-25pt}
		\caption{3 steps ahead forecast}
	\end{subfigure}
	\begin{subfigure}[t]{0.49\textwidth}
		\centering
		\includegraphics[height=0.95\textwidth, width=1.25\textwidth]{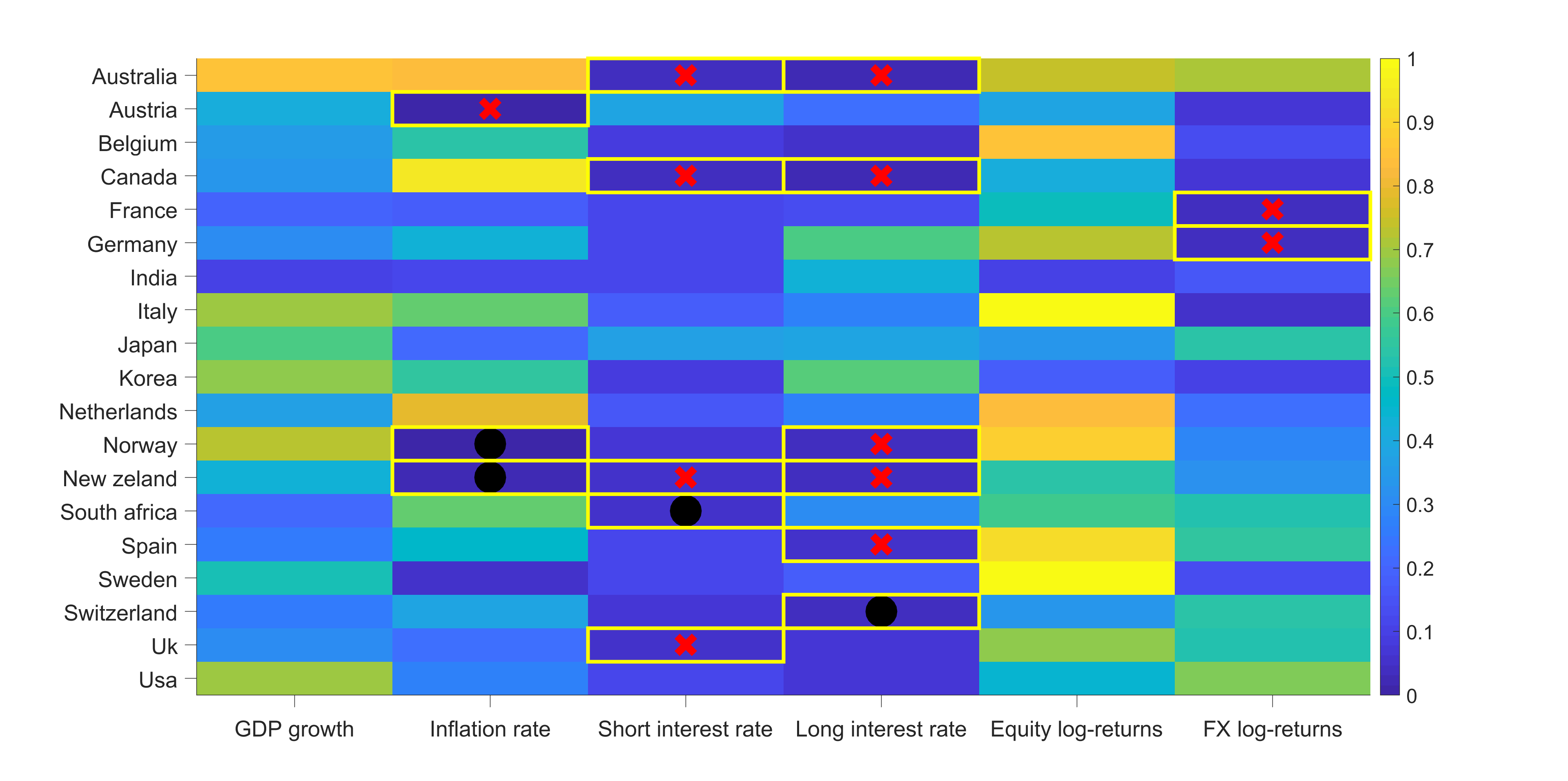}				
		\vspace{-25pt}
		\caption{4 steps ahead forecast}
		
		\hspace*{-1.5cm}
	\end{subfigure}
	\vspace{4pt}
	\caption{ P-values of the DM test between TAR\{1; [1, 1, 1, 1]\} and the VAR(1). Yellow rectangles denote p-values lower than 0.05. Red crosses (\textcolor{red}{$\mathbf{\times}$}) refer to cases in favour of the TAR model while black circles  (\tikzcircle{3pt}) refer to cases in favour of the VAR.}
	\label{fig:fig5}		
	
	\vspace{10pt}
\end{figure}

	In this section, we compare the forecasting performance of the TAR with respect to the VAR model. For this purpose we test the two models performance with the use of the  Diebold-Mariano (DM) test \cite{diebold1995} over four forecasting steps (four quarters ahead). The Diebold-Mariano test is performed between the forecast errors of the TAR and VAR to asses if there is statistical difference in terms of forecasting between the two models. The data is split in 3 samples. The training set corresponds to 70\% of the sample, 20\% is devoted to the parameters optimization for the forecasting model (optimization sample) and the final 10\% of the sample (test sample) is used to compute the forecast errors. For this empirical study, we use as baseline VAR model, a VAR(1) which is estimated for each country separately. For the TAR, we used a 1 lag specification and the optimal Tucker rank $\T{R}_{\bullet}$ and shrinkage parameter $\lambda$ are found in the parameters optimization step. In particular, we estimated the model parameter for a grid of values of the Tucker rank $\T{R}_{\bullet}$ and shrinkage parameter $\lambda$ and then used the optimization sample to infer which specification had better performance. The parameters values we used are:
	\begin{itemize}
	\item[-] Tucker rank $\T{R_{\bullet}}=[f, g, f, g]$ with $f=1,\dots,6$ and $g=1,\dots,19$.
	\item[-] Shrinkage parameter $\lambda=0,0.5,1,2.5, 5$.
	\end{itemize}
	Table \ref{tab3} shows the results of the analysis. We can observe that the BIC computed on the optimization sample is minimized for the specification  $\T{R_{\bullet}}=[1, 1, 1, 1]$ and  $\lambda=5$. This correspondS to a one-factor model with a strong shrinkage and can be attributed to the fact that there is a strong latent driving force within and between dimensions. 
		\begin{table}[h!]
		\begin{center}
			\begin{tabular}{|l|c|c|c|}
				\hline
 
				& \multicolumn{1}{c|}{BIC} & \multicolumn{1}{c|}{$\hat{f}$} & \multicolumn{1}{c|}{$\hat{g}$} \\ \hhline{|=|=|=|=|}
				$\lambda=0$       &            -597.82  &           1                   &        1                       \\ \hline
				$\lambda=0.5$     &            -777.22  &           1                   &        1             \\ \hline
				$\lambda=1$       &            -788.94  &           1                   &        1                     \\ \hline
				$\lambda=2.5$     &            -807.13  &           1                   &        1                     \\ \hline
				$\lambda=5$       &   \textbf{-822.79}  &           1                   &        1                       \\ \hline
				
			\end{tabular}
		\vspace{5pt}
			\caption{Table for the BIC computed on predicted values for the optimization sample.}
			\label{tab3}
			\vspace{-10pt}

		\end{center}
	\end{table}	
	For the computation of the forecasting errors, we use the forecasting recursion:	
	\[\widehat{Y}_{t+h,i}=\chi_i(\widehat{Y}_{t+h-1,i},\Theta_i )\]	
	where $h$ is the forecasting horizon, $\chi_i$ is model $i$ used to forecast and $\Theta_i$ is the set of parameters of model $i$. We then compute the forecast errors for model $i$ as
	 \[FE_{h,i}=Y_{t+h,i}-\widehat{Y}_{t+h,i}\] 
	 and compute the Diebold-Mariano test under the assumption null hypothesis that $\mathbb{E}[FE_{h,1}-FE_{h,2}]=0$ against the alternative  $\mathbb{E}[FE_{h,1}-FE_{h,2}]\neq0$. We refer to the original paper for the technical details of the test \cite{diebold1995}. We show the results of the Diebold-Mariano test for 1, 2, 3 and 4 steps ahead forecasts in Fig. \ref{fig:fig4} and \ref{fig:fig5}. These figures show the p-values associated to the modified Diebold-Mariano test proposed in \cite{harvey97}.\footnote{Their test corrects the standard DM test in order to be robustly used when the number of forecasting errors is small.} As explained above, the null hypothesis states that the two competing models have the same predictive power against the alternative hypothesis for which they have different forecasting accuracy. Values lower than $\alpha=0.05$ represent a rejection of the null hypothesis of equal predictive ability. It is possible to notice that the two models have a similar predictive ability in the majority of the cases. Exceptions to this behaviour are in favour of the TAR. Indeed, roughly 70\% of the rejections come from a lower RMSFE of the TAR model.

We finally analyse the dependency structure over the different modes of the regression residuals. To do so, we compute the covariance over different modes using the flip-flop algorithm proposed in \cite{hoff2011separable}. The procedure is described in Algorithm \ref{f_f} below. This algorithm is based on the assumption of the array Normal distribution for the residual, assumption generally satisfied for macroeconomic data.

\begin{figure}[ht!]
	\captionsetup[subfigure]{justification=centering}
	\begin{subfigure}[t]{0.5\textwidth}
		\includegraphics[height=0.9\textwidth, width=1\textwidth]{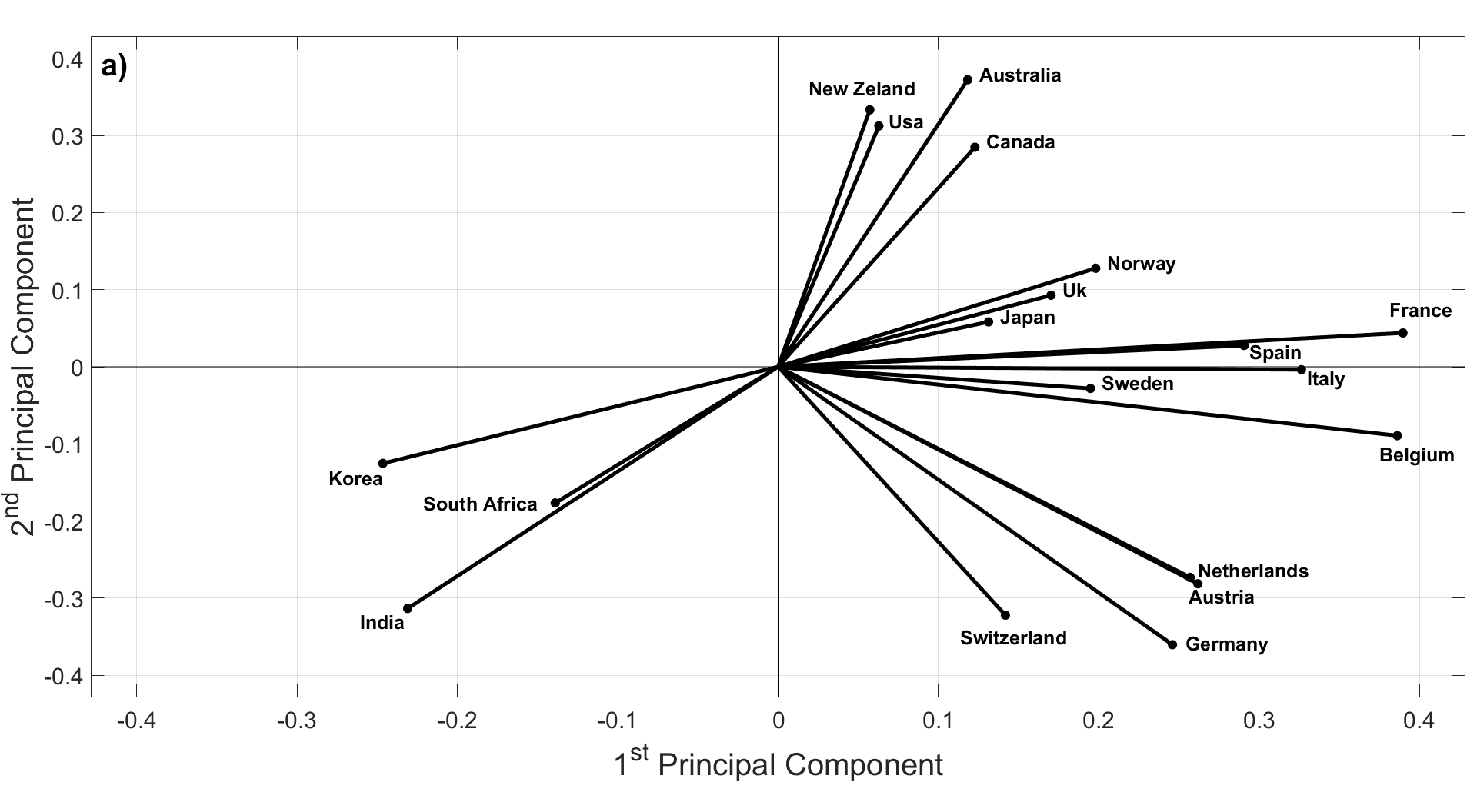}
	\end{subfigure}
	\begin{subfigure}[t]{0.5\textwidth}
		\includegraphics[height=0.9\textwidth, width=1\textwidth]{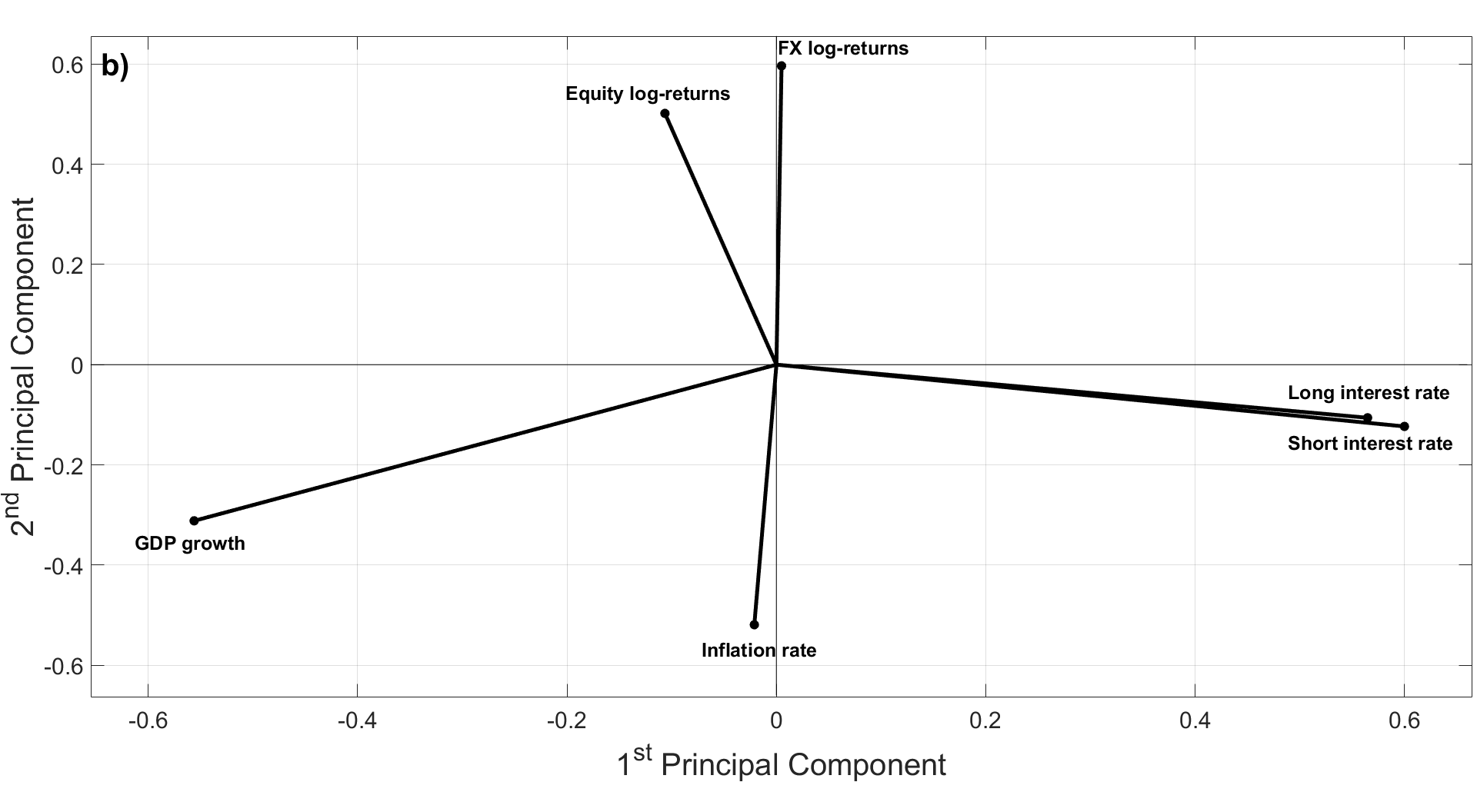}
		\hspace*{-1.5cm}
	\end{subfigure}
	\vspace{4pt}
	\caption{Biplot of the first and second PCA components of the correlation matrices. The subplot a) corresponds to countries while b) to the variables in the dataset.}
	\label{fig:fig3}		
	\vspace{5pt}
\end{figure}

\begin{algorithm}[H]
	\caption{Flip-flop algorithm}\label{f_f}
	\begin{algorithmic}[1]
		\State \textbf{Initialize the algorithm to some $\M{\Sigma}_{1} \cdots \M{\Sigma}_{M+1}$}
			\State Compute $\T{E}=\T{Y}-\T{\widehat{A}}- \langle \T{X},\T{\widehat{B}}  \rangle_{(\T{I_x};\T{I_B})}$

		\State Compute $\T{E}^{(m)}=\T{E}\times_{1} \M{\Sigma}_{1}^{\frac{1}{2}} \cdots \times_{m-1} \M{\Sigma}_{m-1}^{\frac{1}{2}} \times_{m} \M{I}_{m} \times_{m+1} \M{\Sigma}_{m+1}^{\frac{1}{2}} \cdots \times_{M+1} \M{\Sigma}_{M+1}^{\frac{1}{2}} $

		\State Compute $\M{\widehat{\Sigma}}_{m}=\mathbb{E}[\M{E}_{(m)}\M{E}_{(m)}^T]$
		
		\State \textbf{Return  $\widehat{\M{\Sigma}}_{1} \cdots \widehat{\M{\Sigma}}_{M+1}$ }
	\end{algorithmic}
\end{algorithm}
However, $\M{\widehat{\Sigma}}_{m}$ are not identifiable because if we multiply one of the covariance matrices for a scalar $w$ and another covariance matrix for the inverse value $w^{-1}$, the optimization algorithm reaches the same value. For this reason, we work with the correlation matrices rather than with covariance matrices. To analyse the dependency structure between the modes, we perform a Principal Component Analysis PCA on the correlation matrices and plot the first and second components in a biplot. Results are presented in Fig. \ref{fig:fig3}.

As it is possible to notice from the results, we can easily cluster the countries in 5 groups. We have the group of emerging countries (Korea, India and South Africa), the group of Pacific countries (Australia, Canada, New Zealand and USA), the group of strong non-Euro countries with stable inflation (Japan, Norway and UK), the group of European countries with high public debt (Belgium, France, Italy, Sweden, Spain) and the group of strong European Economies with a strong German business culture (Austria, Germany, Netherlands and Switzerland). Regarding the variables, as we could expect, the two interest rates variables are closely related. The FX log-returns and the Equity log-returns are also strongly related since both come from the financial system. Finally, GDP growth and inflation rate are in the same quadrant, making them related in the sense that they both come from the real economy but depicts different aspects of economy growth. These results highlight the good inferential ability of the model and that it can be used also as an inferential tool in addition to as a forecasting model.

	\section{Conclusions}\label{sec_c}
	Multidimensional data constitute an environment where the interrelations are abundant but are too often neglected in empirical works by oversimplifying datasets structure in order to use standard matrix/vector methods. These deconstruction techniques do not only pose a conceptual problem by destroying the real inter-dependencies among dimensions, but they do so also from a computational point of view as the number of parameters to be estimated increases considerably for these models. To overcome these pitfalls, in this paper we have introduced a tensor (Auto)regression (TAR) learning model in which both the dependent and independent variables are tensors. To estimate the huge dimensional coefficient tensor, a Tucker structure is imposed to handle big (possibly dimensionally skewed) data. A penalized tensor regression is also implemented to deal with overfitting and collinear data. Since the model's parameters cannot be estimated all at once, an ALS algorithm is proposed.  To test its reliability and robustness, the model has been tested with two different simulation experiments. The simulation results are robust and they confirm the ability of the model to deal with structured tensor coefficients and with collinearity. The TAR model is then applied to the Foursquare spatio-temporal dataset and to a panel of macroeconomic time series. The forecasting ability of the model is tested against two different leading models: the ALTO model developed in \cite{yu2015} and the Vector Autoregression \cite{var_book}. Results show that the TAR model outperforms the ALTO model applied to the Foursquare spatio-temporal dataset over all the possible scenarios considered. The application to the Macroeconomic data confirms that, even if in the majority of the cases the TAR and VAR models have the same statistical accuracy in forecasting, the proposed model has a better performance when a statistical difference being detected is roughly 70\% of the cases. These results certify the robust nature of the predictions the model is able to perform. We have also analysed the dependency structure over different dimensions of the tensor, finding interesting results in the clustered nature of the data over both the variable and countries dimensions. The model is general enough to be used for different applications. Indeed, this model can be applied to both multivariate and multilinear data in both cross-section and time series domains. This model can be successfully used to estimate and forecast spatio-temporal data, such as the weather, recommendation systems and the type of research problems in the multi-task learning domain. Furthermore, the use of multilinear models for multidimensional data is of key importance for future research in multidimensional big data environments in which the interconnections are sometimes more important than the intra-connections in the dataset. The proposed model goes in this specific direction. Nevertheless, there are still open research questions related to the use of such models. In particular, it is possible to impose a structure on the coefficient tensor, leading to the Structural Tensor Autoregression (STAR) model. This would be useful for both inference and forecasting when information on the structure of $\T{B}$ comes from a theoretical motivation. Let us also mention a related research topic being the use local tensor projection to implement Impulse Response Functions (IRF) and analyse the effect and propagation of shocks from one variable (and possibly country) to the whole system. Finally, since tensor regression can be seen as a generalization of the Vector Error Correction Model (VECM) \cite{vecm}, it would be of considerable research interest the generalization of the VECM specification to the tensor context and analyse its performance. 

	\bibliography{TAR_paper}

\end{document}